\documentclass{article}

\usepackage{amsmath,graphicx,algorithm,algorithmic,xspace,amsfonts,amsthm}
\usepackage{amssymb}
\usepackage{amsthm}
\usepackage{subfigure}
\usepackage{multicol}
\usepackage{tikz}
\usepackage{cite}
\usepackage{verbatim,spconf}

\begin{document}
%
\title{
Bayesian Learning for \\ Low-rank Matrix Reconstruction}
%
%
%

\name{Martin~Sundin,~
		Cristian~R.~Rojas,~
        ~Magnus~Jansson~
		and~Saikat~Chatterjee
\thanks{This work was supported in part by the Swedish Research Council under contract 621-2011-5847.}
\thanks{Parts of this paper was presented at the International Conference on Signal Processing and Communications (SPCOM), July 2014, Bangalore, \mbox{India} \cite{spcom}.}
}
\address{ACCESS Linnaeus Center, School of Electrical Engineering\\
KTH Royal Institute of Technology, Stockholm, Sweden\\
{\small \tt masundi@kth.se, crro@kth.se, janssonm@kth.se, sach@kth.se}}



\maketitle

\begin{abstract}

We develop latent variable models for Bayesian learning based low-rank matrix completion and reconstruction from linear measurements. For under-determined systems, the developed  methods are shown to reconstruct low-rank matrices when neither the rank nor the noise power is known a-priori. We derive relations between the latent variable models and several low-rank promoting penalty functions. The relations justify the use of Kronecker structured covariance matrices in a Gaussian based prior. 
In the methods, we use evidence approximation and expectation-maximization to learn the model parameters. The performance of the methods is evaluated through extensive numerical simulations.



\end{abstract}


%

\section{Introduction}
\label{sec:intro}

Reconstruction of a high dimensional low-rank matrix from a low dimensional measurement vector is a challenging problem.
The low-rank matrix reconstruction (LRMR) problem is inherently under-determined and have been receiving \mbox{considerable} attention \cite{Candes2010noise,Candes2009}
due to its generality over popular sparse reconstruction problems along with many application scopes \cite{Candes2009,Candes2010noise,Fazel2002,Zachariah2012,Fan,yu}. 
Here we consider the LRMR system model 
\begin{align}
\label{eq:measurements}
\mathbf{y} = \mathbf{A}\mathrm{vec}(\mathbf{X})  + \mathbf{n}
\end{align}
where $\mathbf{y} \in \mathbb{R}^m$ is the measurement vector, $\mathbf{A} \in \mathbb{R}^{m \times pq}$ is the linear measurement matrix, $\mathbf{X}\in \mathbb{R}^{p \times q}$ is the low-rank matrix, $\mathbf{n}\in\mathbb{R}^m$ is additive noise (typically assumed to be zero-mean Gaussian with covariance $\mathrm{Cov}(\mathbf{n}) = \beta^{-1} \mathbf{I}_m$) and $\mathrm{vec}(\cdot)$ is the vectorization operator. With $m < pq$, the setup is underdetermined and the task is the reconstruction (or estimation) of $\mathbf{X}$ from $\mathbf{y}$.
To deal with the underdetermined setup, a typical and much used strategy is to use a regularization in the 
 reconstruction
cost function. Regularization brings in the information about low rank priors.
A typical type I estimator is
\begin{align}
\label{eq:marginal_MAP}
\hat{\mathbf{X}} = \arg \min_{\mathbf{X}} \beta ||\mathbf{y - A}\mathrm{vec}(\mathbf{X})||_2^2 + g(\mathbf{X}) , 
\end{align}
where $\beta > 0$ is a regularization parameter and $g(\cdot)$ is a fixed penalty function that promotes low rank in $\hat{\mathbf{X}}$.
Common low-rank penalties in the literature \cite{Fazel2002,Candes2010noise,irls} are
\begin{align}
&g(\mathbf{X}) = ||\mathbf{X}||_* = \mathrm{tr}( (\mathbf{X X^\top})^{1/2} ) , \tag{nuclear norm}\\
&g(\mathbf{X}) =  \mathrm{tr}( (\mathbf{X X^\top})^{s/2} ) , \tag{Schatten s-norm}\\
&g(\mathbf{X}) = \log |\mathbf{X X^\top} + \epsilon \mathbf{I}_p | , \tag{log-determinant penalty}
\end{align}
where $\mathrm{tr}(\cdot)$ denotes the matrix trace, $|\cdot |$ denotes determinant, and $0< s \leq1$ and $\epsilon >0$.
We mention that the nuclear norm penalty is a convex function.

In the literature, LRMR algorithms can be categorized in three types: convex optimization 
\cite{Fazel2002,Candes2009,Candes2010noise,Candes10,Cai2010,Boyd}, greedy solutions \cite{Admira,Tang2011,Zachariah2012, Johnstone, irls} and 
Bayesian learning \cite{Babacan}. Most of these existing algorithms are highly motivated from analogous 
algorithms used for standard sparse reconstruction problems, such as compressed sensing where 
$\mathrm{vec}(\mathbf{X})$ in \eqref{eq:measurements} is replaced by a sparse vector. 
Using convex optimization we can solve \eqref{eq:marginal_MAP} when $g(\mathbf{X})$ is the nuclear norm,
which is an analogue of using $\ell_1$-norm in sparse reconstruction problems.
Further, greedy algorithms, such as iteratively reweighted least squares \cite{irls} solves \eqref{eq:marginal_MAP}
by using algebraic approximations.
While convex optimization and greedy solutions are popular they often need more a-priori information than 
knowledge about structure of the signal under reconstruction; for example, convex optimization algorithms
need information about the strength of the measurement noise to fix the parameter $\beta$, and greedy algorithms need information about rank. 
In absence of such a-priori information, Bayesian learning is a preferred strategy to use.
Bayesian learning is capable of estimating the necessary information from measurement data. 
In Bayesian learning we evaluate the posterior $p(\mathbf{X}|\mathbf{y})$ with the knowledge of prior $p(\mathbf{X})$.
If $\mathbf{X}$ has a prior distribution $p(\mathbf{X}) \propto e^{-\frac{1}{2}g(\mathbf{X})}$ 
and the noise is distributed as $\mathbf{n} \sim \mathcal{N}(\mathbf{0},\beta^{-1} \mathbf{I}_m)$, then the maximum-a-posteriori (MAP)
estimate can be interpreted as the type I estimate in \eqref{eq:marginal_MAP}.
As type I estimation requires more information (such as $\beta$), type II estimators are often more useful.
Type II estimation techniques use hyper-parameters in the form of latent variables with prior distributions.
While for sparse reconstruction problems, Bayesian learning via type II estimation
in the form of relevance vector machine \cite{Tipping1,Tipping2} and sparse Bayesian learning \cite{zhang,zhang2} have gained significant popularity, the endeavor to design type II estimation algorithms for LRMR is found to be limited. 
In \cite{Babacan}, direct use of sparse Bayesian learning was used to realize an LRMR reconstruction algorithm.
Bayesian approaches were used in \cite{Carin,wipf_lowrank} for a problem setup with a combination of 
low rank and sparse priors, called principal component pursuit \cite{candes2011robust}. In \cite{Carin}, Gaussian and Bernoulli variables was used and the parameters were estimated using Markov Chain Monte Carlo while in \cite{Wipf} an empirical Bayesian approach was used.
Type II estimation methods are typically iterative where latent variables are usually treated via variational techniques \cite{wipf_lowrank,Babacan}, 
evidence approximation \cite{Tipping1,Tipping2}, expectation maximization \cite{zhang,zhang2} and Markov chain Monte Carlo \cite{Carin,yu}.

Our objective in this paper is to develop new type II estimation methods for LRMR.
Borrowing ideas from type II estimation techniques for sparse reconstruction, such as
the relevance vector machine and sparse Bayesian learning algorithms, we model a low-rank matrix
by a multiplication of precision matrices and an i.i.d. Gaussian matrix. The use of precision
matrices helps to realize low-rank structures. The precision
matrices are characterized by hyper-parameters which are treated as latent variables.
The main contributions of this paper are as follows.
\begin{enumerate}
\item We introduce one-sided and two-sided precision matrix based models.
\item We show how the Schatten s-norm and log-determinant penalty functions  
are related to latent variable models in the sense of MAP estimation via type I estimator \eqref{eq:marginal_MAP}.
\item For all new type II estimation methods, we derive update equations for all parameters in iterations.
The methods are based on evidence approximation and expectation-maximization.
\item The methods are compared numerically to existing methods, such as the 
Bayesian learning method of \cite{Babacan} and nuclear norm based convex optimization method \cite{Candes2010noise}.
\end{enumerate}

We are aware that evidence approximation and expectation-maximization are unable to provide globally optimal solutions.
Hence we are unable to provide performance guarantees for our methods.
This paper is organized as follows. We discuss the preliminaries of sparse Bayesian learning in section~\ref{sec:preliminaries}. In section~\ref{sec:one_sided} we introduce one-sided precisions for matrices and derive the relations to type I estimators. Two-sided precisions are introduced in section~\ref{sec:two_sided} and in section~\ref{sec:algorithms} we derive the update equations for the parameters. In section~\ref{sec:simulations} we numerically compare the performance of the algorithms for matrix reconstruction and matrix completion.

\subsection{Preliminaries}
\label{sec:preliminaries}

In this section, we explain the relevance vector machine (RVM) \cite{Tipping1,Tipping2,bayesianCS} and sparse Bayesian learning (SBL) 
methods \cite{zhang,zhang2,Wipf} for a standard sparse reconstruction problem. The setup is
\begin{eqnarray}
\mathbf{y = Ax + n},
\label{eq:CS_setup}
\end{eqnarray}
where $\mathbf{x}\in\mathbb{R}^n$ is the sparse vector to be reconstructed from the measurement vector $\mathbf{y}\in\mathbb{R}^m$ and
$\mathbf{n} \sim \mathcal{N}(\mathbf{0},\beta^{-1} \mathbf{I}_m)$ is the additive measurement noise.
The approach is to model the sparse signal $\mathbf{x} = [x_1 \, x_2 \, \hdots \, x_n]^{\top}$ as
\begin{align}
\label{eq:vec_model}
x_i = \gamma_i^{-1/2} u_i , 
\end{align}
where $u_i \sim \mathcal{N}(0,1)$ and $\gamma_i > 0$ is the precision of $x_i$. This is equivalent to setting
\begin{align*}
p(x_i | \gamma_i) = \sqrt{\frac{\gamma_i}{2\pi}} \exp (-\gamma_i x_i^2 /2) = \mathcal{N}(x_i |0,\gamma_i^{-1}).
\end{align*}
The main idea is to use a learning algorithm for which several precisions go to infinity,
leading to sparse reconstruction. Alternatively said, the use of precisions allows to inculcate
\emph{dominance} of a few components over other components in a sparse vector.
Note that $\boldsymbol{\gamma} = [\gamma_1 \, \gamma_2 \, \hdots \, \gamma_n]^{\top}$ and $\beta$ are latent variables that also need to be estimated. 
We find the posterior
\begin{align*}
p(\mathbf{x}|\mathbf{y}) = \int p(\mathbf{x}|\mathbf{y},\boldsymbol{\gamma},\beta)  \, p(\boldsymbol{\gamma},\beta | \mathbf{y}) \, d\boldsymbol{\gamma} \, d\beta 
 \approx p(\mathbf{x}|\mathbf{y},\hat{\boldsymbol{\gamma}},\hat{\beta}), \nonumber
 \end{align*} 
if $p(\boldsymbol{\gamma},\beta | \mathbf{y})$ is assumed sharply peaked around $\hat{\boldsymbol{\gamma}}$, $\hat{\beta}$ (this is version of the so-called Laplace approximation described in Appendix \ref{appendix:laplace}). 
Assuming the knowledge of $\boldsymbol{\gamma} = \hat{\boldsymbol{\gamma}}$ and $\beta=\hat{\beta}$, the MAP estimate is
\begin{eqnarray}
\hat{\mathbf{x}} \triangleq [\hat{x}_1 \, \hat{x}_2 \, \hdots \, \hat{x}_n]^{\top} \leftarrow \arg \max_{\mathbf{x}} p(\mathbf{x}|\mathbf{y}, \boldsymbol{\gamma}, \beta) = \beta \boldsymbol{\Sigma} \mathbf{A}^\top \mathbf{y}
\label{eq:MAP_x_rvm}
\end{eqnarray}
where $\boldsymbol{\Sigma} = (\mathrm{diag}(\boldsymbol{\gamma}) + \beta \mathbf{A^\top A} )^{-1}$. In the notion of iterative updates, we use $\leftarrow$ to denote the assignment operator.
The precisions $\boldsymbol{\gamma}$ and $\beta$ are estimated by
\begin{align*}
(\gamma_i^{new},\beta^{new}) & \leftarrow  \arg \max_{\gamma_i,\beta} p(\mathbf{y},\boldsymbol{\gamma},\beta) \\
& = \arg \max_{\gamma_i,\beta} p(\mathbf{y}|\boldsymbol{\gamma},\beta) \, p(\boldsymbol{\gamma}) \, p(\beta),     
\end{align*}
where $p(\boldsymbol{\gamma}) = \prod_i p(\gamma_i)$ and $p(\mathbf{y}|\boldsymbol{\gamma},\beta) = \mathcal{N}(\mathbf{y}|\mathbf{0},(\mathbf{A}\mathrm{diag}(\boldsymbol{\gamma}) \mathbf{A}^\top + \beta^{-1} \mathbf{I}_m)^{-1})$. Gamma distributions are typically chosen as hyper-priors for $p(\gamma_i)$ and $p(\beta)$ with the form
\begin{align*}
p(\beta) = \mathrm{Gamma}(\beta | a+1,b) = \frac{b^{a+1}}{\Gamma (a+1)} \beta^a e^{-b\beta}, 
\end{align*}
with $a > -1$, $b>0$ and $\beta \geq 0$. The evaluation of $(\gamma_i^{new},\beta^{new})$ leads to coupled equations and are therefore solved approximately as
\begin{eqnarray}
\gamma_i \leftarrow  \frac{1 - \gamma_i \Sigma_{ii} + 2a}{\hat{x}_i^2 + 2b} \,\, \mathrm{and} \,\,
\beta\leftarrow  \frac{\sum_{i=1}^n \gamma_i \Sigma_{ii} + 2a}{||\mathbf{y - A\hat{x}}||_2^2 + 2b}, 
\label{eq:EdidenceApprox_rvm}
\end{eqnarray}
where $\Sigma_{ii}$ is the $i$'th diagonal element of $\boldsymbol{\Sigma}$. The parameters of the Gamma distributions for 
$p(\boldsymbol{\gamma})$ and $p(\beta)$ are typically chosen to be non-informative, i.e. $a,b \to 0$.
The update solutions of \eqref{eq:MAP_x_rvm} and \eqref{eq:EdidenceApprox_rvm} are repeated iteratively until convergence.
In sparse Bayesian learning algorithm a standard expectation-maximization framework is used 
to estimate $\boldsymbol{\gamma}$ and $\beta$.

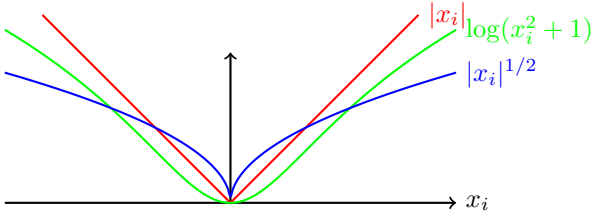
\begin{figure}
\begin{center}
\begin{tikzpicture}[domain=0:4]
	\draw[thick,->] (-3,0) -- (3,0) node[right] {$x_i$};
	\draw[thick,->] (0,0) -- (0,2);
	\draw[color=red,thick] (-2.5,2.5) -- (0,0) -- (2.5,2.5) node[right] {$|x_i|$};
	\draw[domain=-3:3,smooth,samples=300,variable=\x,color=green,thick] plot ({\x},{ln(\x*\x + 1)}) node[right] {$\log (x_i^2 + 1)$};
	\draw[domain=-3:3,smooth,samples=300,variable=\x,color=blue,thick] plot ({\x},{sqrt(abs(\x))}) node[right] {$|x_i|^{1/2}$};
\end{tikzpicture}
\end{center}
\caption{Illustration of different sparsity promoting penalty functions.}
\label{Fig:illustration}
\end{figure}

Finally we mention that the RVM and SBL methods have connection with type I estimation \cite{Wipf,Babacan_laplacian}.
If the precisions have arbitrary prior distributions $p(\gamma_i)$, then the marginal distribution of $x_i$ becomes
\begin{align*}
p(x_i) = \int p(x_i | \gamma_i) \, p(\gamma_i) \, d\gamma_i \propto e^{-h(x_i)/2} ,
\end{align*}
for some function $h(\cdot)$.
Given \eqref{eq:CS_setup} and for a known $\beta$,  the MAP estimate is 
\begin{align*}
\hat{\mathbf{x}} = \arg \min_\mathbf{x} \left( \beta || \mathbf{y - Ax} ||_2^2 + \sum_{i=1}^n h(x_i) \right) . 
\end{align*}
If $p(\gamma_i)$ is a gamma prior then $p(x_i)$ is a Student-$t$ distribution with 
$h(x_i) = \mathrm{constant} \times \log (x_i^2 + \mathrm{constant}) $.
One rule of thumb is that a "more" concave $h(x_i)$ gives a more sparsity promoting model \cite{Wipf}, see some example functions in Figure~\ref{Fig:illustration}. In the figure, $h(x_i) = |x_i|$ corresponds to a Laplace distributed variable $x_i$, $\log(x_i^2+1)$ to a Student-$t$ and $|x_i|^{1/2}$ to a generalized normal distribution \cite{general_gaussian}. 
The relation between the sparsity promoting penalty function $h(x_i)$ and the corresponding
prior $p(\gamma_i)$ of the latent variable $\gamma_i$ was discussed in \cite{Wipf}, see also \cite{Rojas} and
\cite{Stoica2014}.

\section{One-sided precision based model}

\label{sec:one_sided}

The structure of a low-rank matrix $\mathbf{X}$ is characterized by the dominant singular vectors and singular values.
Like the use of precisions in \eqref{eq:vec_model} for the standard sparse reconstruction problem via inculcating dominance,
we propose to model the low-rank matrix $\mathbf{X}$ as
\begin{align}
\mathbf{X} = \boldsymbol{\alpha}^{-1/2} \mathbf{U}, 
\label{eq:X_oneprior_1}
\end{align}
where the components of $\mathbf{U} \in \mathbb{R}^{p \times q}$ are i.i.d. $\mathcal{N}(0,1)$ and $\boldsymbol{\alpha} \in \mathbb{R}^{p \times p}$ is a positive definite random matrix (which distribution will be described later). This is equivalent to 
\begin{align}
p(\mathbf{X} | \boldsymbol{\alpha}) = \frac{|\boldsymbol{\alpha}|^{q/2}}{(2\pi)^{pq/2}} \exp \left( - \frac{1}{2} \mathrm{tr}(\mathbf{X}^\top \boldsymbol{\alpha} \mathbf{X}) \right). 
\label{eq:X_oneprior}
\end{align}
Denoting $\mathbf{Z = XX^\top}$ and $\mathrm{tr}(\mathbf{X}^\top \boldsymbol{\alpha} \mathbf{X}) = \mathrm{tr}(\boldsymbol{\alpha} \mathbf{X} \mathbf{X}^\top) = \mathrm{tr}(\boldsymbol{\alpha} \mathbf{Z}) $, we evaluate
\begin{align}
\label{eq:marginal_x}
p(\mathbf{X}) & = \int_{\boldsymbol{\alpha} \succ \mathbf{0}} p(\mathbf{X} | \boldsymbol{\alpha})  \, p(\boldsymbol{\alpha}) \, d \boldsymbol{\alpha} \nonumber \\
 & = \int_{\boldsymbol{\alpha} \succ \mathbf{0}} e^{-\frac{1}{2} \mathrm{tr}(\boldsymbol{\alpha}\mathbf{Z})} \frac{|\boldsymbol{\alpha}|^{q/2}}{(2\pi)^{pq/2}}  p(\boldsymbol{\alpha}) d \boldsymbol{\alpha}  \\
 & \propto e^{- \frac{1}{2} \tilde{g}(\mathbf{Z})} \propto e^{-\frac{1}{2} g(\mathbf{X})}. \nonumber
\end{align}
We note that $g(\mathbf{X})$ must have the special form $g(\mathbf{X}) = \tilde{g}(\mathbf{XX^\top})$ for $p(\mathbf{X})$ \eqref{eq:marginal_x} to hold (as $\boldsymbol{\alpha}$ is integrated out).
As $p(\mathbf{X}) \propto e^{-\frac{1}{2} g(\mathbf{X})}$, the resulting MAP estimator can be interpreted as the type I estimator \eqref{eq:marginal_MAP}. 

\subsection{Relation between priors}

Next we investigate the relation between the priors $p(\mathbf{X})$ and $p(\boldsymbol{\alpha})$.
The motivation is that the relations are necessary for designing practical learning algorithms. 
From \eqref{eq:marginal_x}, we note that $p(\mathbf{X})$ is the Laplace transform of $|\boldsymbol{\alpha}|^{q/2} p(\boldsymbol{\alpha})/(2\pi)^{pq/2}$ \cite{Terras}, which establishes the relation. Naturally, we can find $p(\boldsymbol{\alpha})$
by the inverse Laplace transform \cite{Terras} as follows
\begin{align}
\label{eq:inverse_laplace}
p(\boldsymbol{\alpha}) \propto |\boldsymbol{\alpha}|^{-q/2} \int\limits_{\mathrm{Re} \, \mathbf{Z} = \boldsymbol{\alpha}_* } e^{\frac{1}{2} \mathrm{tr}(\boldsymbol{\alpha}\mathbf{Z})}  e^{-\frac{1}{2} \tilde{g}(\mathbf{Z})} d\mathbf{Z} , 
\end{align}
where the integral is taken over all symmetric matrices $\mathbf{Z} \in \mathbb{C}^{p\times p}$ such that $\mathrm{Re} \, \mathbf{Z} = \boldsymbol{\alpha}_*$ where $\boldsymbol{\alpha}_*$ is a real matrix so that the contour path of integration is in the region of convergence of the integrand.
While the Laplace transform characterizes the exact relation between priors, the computation is non-trivial and often analytically intractable.
In practice, a standard approach is to use the Laplace approximation \cite{mackay} where typically the mode of the distribution under approximation is found first and then a Gaussian distribution is modeled around that mode. 
Let $p(\boldsymbol{\alpha})$ have the form $p(\boldsymbol{\alpha}) \propto e^{-\frac{1}{2} K(\boldsymbol{\alpha})}$; then
the Laplace approximation becomes
\begin{align*}
\tilde{g}(\mathbf{Z}) = & \min_{\boldsymbol{\alpha} \succ \mathbf{0}} \left\{ \mathrm{tr}(\boldsymbol{\alpha}\mathbf{Z}) - q\log |\boldsymbol{\alpha}| + K(\boldsymbol{\alpha}) \right\} \\
& - \log \left| \mathbf{H} \right| + \mathrm{constant}, 
\end{align*}
where $\mathbf{H}$ is the Hessian of $\mathrm{tr}(\boldsymbol{\alpha}\mathbf{Z}) - q\log |\boldsymbol{\alpha}| + K(\boldsymbol{\alpha})$ 
evaluated at the minima (which is assumed to exist). 
The derivation of the Laplace approximation is shown in Appendix \ref{appendix:laplace}.

Denoting $\tilde{K}(\boldsymbol{\alpha}) = q\log |\boldsymbol{\alpha}| - K(\boldsymbol{\alpha})$ and assuming that the Hessian is constant (independent of $\mathbf{Z}$) we get that
\begin{align*}
\tilde{g}(\mathbf{Z}) = \min_{\boldsymbol{\alpha} \succ \mathbf{0}} \left\{ \mathrm{tr}(\boldsymbol{\alpha}\mathbf{Z}) - \tilde{K}(\boldsymbol{\alpha}) \right\}, 
\end{align*}
where we absorbed the constants terms into the normalization factor of $p(\mathbf{X})$.
We find that $\tilde{g}(\boldsymbol{Z})$ is the concave conjugate of $\tilde{K}(\boldsymbol{\alpha})$ \cite{Boyd}.
Hence, for a given $\tilde{g}(\boldsymbol{Z})$ we can recover $\tilde{K}(\boldsymbol{\alpha})$ as
\begin{align}
\label{eq:inverse_conjugate}
\tilde{K}(\boldsymbol{\alpha}) = \min_{\boldsymbol{Z} \succ \mathbf{0}} \left\{ \mathrm{tr}(\boldsymbol{\alpha}\mathbf{Z}) - \tilde{g}(\boldsymbol{Z}) \right\}
\end{align}
if $\tilde{K}(\boldsymbol{\alpha})$ is concave (which holds under the assumption that $K(\boldsymbol{\alpha})$ is convex). Further, we can find $K(\boldsymbol{\alpha})$ from $\tilde{K}(\boldsymbol{\alpha})$
followed by solving the prior $p(\boldsymbol{\alpha}) \propto e^{-\frac{1}{2} K(\boldsymbol{\alpha})}$. Using the concave conjugate relation \eqref{eq:inverse_conjugate},
we now deal with the task of finding appropriate $K(\boldsymbol{\alpha})$ for two example low-rank promoting penalty functions, as follows.

\begin{enumerate}
 \item{\emph{For Schatten $s$-norm:}} The Schatten $s$-norm based penalty function is $g(\mathbf{X}) = \mathrm{tr}((\mathbf{XX^\top})^{s/2})$.  We here use a regularized Schatten $s$-norm based penalty function as 
 \begin{align}
 g(\mathbf{X}) = & \mathrm{tr}((\mathbf{XX^\top + \epsilon \mathbf{I}_p})^{s/2}), 
 \label{eq:Schatten_reg_penalty}
 \end{align}
 where the use of $\epsilon >0$ helps to bring numerical stability to the algorithms in Section~\ref{sec:algorithms}. 
 For the penalty function \eqref{eq:Schatten_reg_penalty}, we find the appropriate $K(\boldsymbol{\alpha})$ as 
 \begin{align}
 K(\boldsymbol{\alpha}) = C_s \, \mathrm{tr}(\boldsymbol{\alpha}^{-\frac{s}{2-s}}) 
 + q\log|\boldsymbol{\alpha}| + \epsilon \, \mathrm{tr}(\boldsymbol{\alpha}), 
 \label{eq:schatten_latent}
 \end{align}
 where $C_s = \frac{2-s}{s} \left( \frac{2}{s}\right)^{-\frac{s}{2-s}}$. The derivation of \eqref{eq:schatten_latent} is given in Appendix \ref{appendix2}.
 Note that, for $s=1$, $g(\mathbf{X})$ becomes the regularized nuclear norm based penalty function
 \begin{align*}
 g(\mathbf{X})  = \mathrm{tr}((\mathbf{XX^\top} + \epsilon \mathbf{I}_p)^{\frac{1}{2}}) = \sum_{i=1}^{\min (p,q)} (\sigma_i(\mathbf{X})^2 + \epsilon)^{\frac{1}{2}}. 
 \end{align*}
 
 \item{\emph{Log-determinant penalty:}} For the log-determinant based penalty function 
 \begin{align}
 g(\mathbf{X}) = \nu \log \left| \mathbf{XX^\top} + \epsilon \mathbf{I}_p \right|,
 \end{align}
 where $\nu > q-2$ is a real number,  we find $K(\boldsymbol{\alpha})$ as 
 \begin{align}
 K(\boldsymbol{\alpha}) = \epsilon \, \mathrm{tr}( \boldsymbol{\alpha}) + \left(q - \nu \right)\log |\boldsymbol{\alpha}|. 
 \label{eq:rsvm_latent}
 \end{align}
 As $p(\boldsymbol{\alpha}) \propto e^{-\frac{1}{2} K(\boldsymbol{\alpha})}$, we find that the prior $\boldsymbol{\alpha}$ is Wishart distributed
 (Wishart is a conjugate prior the distribution \eqref{eq:X_oneprior}).
 For a scalar instead of a matrix, the prior distribution becomes a Gamma distribution as used in the standard RVM and SBL.
\end{enumerate}
 
We have discussed a left-sided precision based model~\eqref{eq:X_oneprior_1} in this section, but the same strategy can be easily extended to form a right-sided precision based model. 
Then a natural question arises, which model to use? Our hypothesis is that the user choice stems from minimizing the number of variables to estimate. If the low-rank matrix is fat 
then the left-sided model should be used, otherwise the right-sided model. 
A further question arises on the prospect of developing a two sided precision based model,
which is described in the next section.

\section{Two-sided precision based model}
\label{sec:two_sided}

In this section, we propose to use precision matrices on both sides to model a random low-rank matrix. 
We call this the two-sided precision based model.
Our hypothesis is that the two-sided precision helps to enhance dominance of a few singular vectors. 
For low-rank modeling, we make the following ansatz
\begin{align}
\mathbf{X} = \boldsymbol{\alpha}_L^{-1/2} \, \mathbf{U} \, \boldsymbol{\alpha}_R^{-1/2}
\label{eq:X_doubleprior_1}
\end{align} 
where $\boldsymbol{\alpha}_L \in \mathbb{R}^{p \times p}$ and $\boldsymbol{\alpha}_R \in \mathbb{R}^{q \times q}$ are positive definite random matrices.
Using the relation $\mathrm{vec}(\mathbf{X}) = (\boldsymbol{\alpha}_R^{-1/2} \otimes \boldsymbol{\alpha}_L^{-1/2}) \, \mathrm{vec}(\mathbf{U}) = 
(\boldsymbol{\alpha}_R \otimes \boldsymbol{\alpha}_L)^{-1/2} \, \mathrm{vec}(\mathbf{U})$, we find
\begin{align}
&p(\mathbf{X} | \boldsymbol{\alpha}_L,\boldsymbol{\alpha}_R) \nonumber\\
&= \frac{| \boldsymbol{\alpha}_R \otimes \boldsymbol{\alpha}_L |^{1/2}}{(2\pi)^{pq/2}} \exp \left( - \mathrm{vec}(\mathbf{X})^\top (\boldsymbol{\alpha}_R \otimes \boldsymbol{\alpha}_L) \mathrm{vec}(\mathbf{X})/2 \right) \nonumber \\
& =\frac{|\boldsymbol{\alpha}_L|^{q/2}|\boldsymbol{\alpha}_R|^{p/2}}{(2\pi)^{pq/2}} \exp \left( - \mathrm{tr}(\mathbf{X}^\top \boldsymbol{\alpha}_L \mathbf{X} \boldsymbol{\alpha}_R)/2\right).
\label{eq:X_doubleprior}
\end{align}
To promote low-rank, we use a prior distribution $p(\boldsymbol{\alpha}_L,\boldsymbol{\alpha}_R) = p(\boldsymbol{\alpha}_L) p(\boldsymbol{\alpha}_R) $. The marginal distribution of $\mathbf{X}$ is 
\begin{align}
p(\mathbf{X}) =  \int\limits_{ \substack{ \boldsymbol{\alpha}_L \succ \mathbf{0} \\ \boldsymbol{\alpha}_R \succ \mathbf{0} }} p(\mathbf{X} | \boldsymbol{\alpha}_L,\boldsymbol{\alpha}_R) \, p(\boldsymbol{\alpha}_L) \, p(\boldsymbol{\alpha}_R) \, d\boldsymbol{\alpha}_R \, d\boldsymbol{\alpha}_L . 
\label{eq:marginal_X_doubleprior}
\end{align}
We have noticed that the use of \eqref{eq:X_doubleprior} in evaluating \eqref{eq:marginal_X_doubleprior} does not bring out suitable 
connections between the resulting $p(\mathbf{X})$ functions and the usual low-rank promoting $g(\mathbf{X})$ functions (such as nuclear norm,
Schatten s-norm and log-determinant). Thus it is non-trivial to establish a direct connection between $p(\mathbf{X} | \boldsymbol{\alpha}_L,\boldsymbol{\alpha}_R)$ of \eqref{eq:X_doubleprior} and the type I estimator of \eqref{eq:marginal_MAP}.

Instead of a direct connection we can establish an indirect connection by an approximation.
For a given $(\boldsymbol{\alpha}_R,\beta)$ and by marginalizing over $\boldsymbol{\alpha}_L$, we have 
$p(\mathbf{X} | \boldsymbol{\alpha}_R ) \propto e^{- \frac{1}{2} \tilde{g}(\mathbf{X} \boldsymbol{\alpha}_R \mathbf{X}^{\top})}$ and hence
the corresponding type I estimator cost function is
\begin{align}
\min_{\mathbf{X}} \beta ||\mathbf{y - A}\mathrm{vec}(\mathbf{X})||_2^2 + \tilde{g}(\mathbf{X} \boldsymbol{\alpha}_R \mathbf{X}^{\top}). 
\end{align}
A similar cost function can be found for a given $(\boldsymbol{\alpha}_L,\beta)$ by marginalizing over $\boldsymbol{\alpha}_R$.
We discuss the roles of $\boldsymbol{\alpha}_L$ and $\boldsymbol{\alpha}_R$ in the next section.

\subsection{Interpretation of the precisions}

From \eqref{eq:X_doubleprior_1}, we can see that the column and row vectors of $\mathbf{X}$ are in the range spaces of $\boldsymbol{\alpha}_L^{-1}$ and
$\boldsymbol{\alpha}_R^{-1}$, respectively. 


Further let us interpret this in a statistical sense with the note that a skewed precision matrix
comprises of correlated components. 
Let us denote  the $(i,j)$th component of $\boldsymbol{\alpha}_R^{-1}$ by $[\boldsymbol{\alpha}_R^{-1}]_{ij}$.
If $\boldsymbol{\alpha}_R^{-1}$ is highly skewed then $[\boldsymbol{\alpha}_R^{-1}]_{ij}$ and $[\boldsymbol{\alpha}_R^{-1}]_{ii}$
are highly correlated. 
Suppose $\mathbf{x}_i \in \mathbb{R}^p$ denotes the $i$'th column vector of $\mathbf{X}$. 
Then following \eqref{eq:X_doubleprior}, we can write 
\begin{align*}
\left[
\begin{array}{c}
 \mathbf{x}_i \\
 \mathbf{x}_j
\end{array}
\right] \sim 
\mathcal{N}
\left(
\left[
\begin{array}{c}
 \mathbf{0} \\
 \mathbf{0}
\end{array}
\right]
,
\left[
\begin{array}{cc}
 [\boldsymbol{\alpha}_R^{-1}]_{ii} \, \boldsymbol{\alpha}_L^{-1} & [\boldsymbol{\alpha}_R^{-1}]_{ij} \, \boldsymbol{\alpha}_L^{-1} \\
 { [\boldsymbol{\alpha}_R^{-1}]_{ji} \, \boldsymbol{\alpha}_L^{-1} } & [\boldsymbol{\alpha}_R^{-1}]_{jj} \, \boldsymbol{\alpha}_L^{-1}
\end{array}
\right]
\right).
\end{align*}
The above relation shows that a presence of highly skewed $\boldsymbol{\alpha}_R^{-1}$
leads to the cross-correlation $[\boldsymbol{\alpha}_R^{-1}]_{ij} \, \boldsymbol{\alpha}_L^{-1}$ 
between $\mathbf{x}_i$ and $\mathbf{x}_j$ that is comparably strong to the auto-correlations 
$[\boldsymbol{\alpha}_R^{-1}]_{ii} \, \boldsymbol{\alpha}_L^{-1}$. We mention that
a low-rank property can be established in a qualitative statistical sense by the presence
of columns having strong cross-correlation. 
The one-sided precision based model
can be seen as the two-sided model where $\boldsymbol{\alpha}_R^{-1} = \mathbf{I}_q$.
Hence the one-sided precision based model is unable to capture information about 
cross-correlation between columns of $\mathbf{X}$. A similar argument can be made for the right sided precision based model where $\boldsymbol{\alpha}_L^{-1} = \mathbf{I}_p$.


\section{Practical algorithms}
\label{sec:algorithms}

Considering the potential of two-sided precision matrices, the optimal inference problem is
\begin{align*}
 \max \, p(\mathbf{X,y}, \boldsymbol{\alpha}_L, \boldsymbol{\alpha}_R, \beta)
\end{align*}
which is the MAP estimator for amenable priors and often connected with the type I estimator in \eqref{eq:marginal_MAP}.
Direct handling of the optimal inference problem is limited due to lack of analytical tractability.   
Therefore various approximations are used to design practical algorithms which are also type II estimators.
This section is dedicated to design new type II estimators via evidence approximation (as used by the RVM) and expectation-maximization (as used in SBL) approaches.

\subsection{Evidence approximation}
In the evidence approximation, we iteratively update the parameters as
\begin{align}
 \begin{aligned}
    \hat{\mathbf{X}} \leftarrow \arg \max_{\mathbf{X}} p(\mathbf{X}| \mathbf{y} , \boldsymbol{\alpha}_L, \boldsymbol{\alpha}_R, \beta) ,
   \end{aligned} \label{eq:LMMSE_estimate_X} \\ 
 \begin{aligned} 
    \beta &\leftarrow \arg \max_{\beta} p(\mathbf{y} , \boldsymbol{\alpha}_L, \boldsymbol{\alpha}_R, \beta) ,
   \end{aligned} \label{eq:RVM_like_estimate_beta} \\
 \left.
 \begin{aligned}
 \boldsymbol{\alpha}_L &\leftarrow \arg \max_{\boldsymbol{\alpha}_L} p(\mathbf{y} , \boldsymbol{\alpha}_L, \boldsymbol{\alpha}_R, \beta) , \\
 \boldsymbol{\alpha}_R &\leftarrow \arg \max_{\boldsymbol{\alpha}_R} p(\mathbf{y} , \boldsymbol{\alpha}_L, \boldsymbol{\alpha}_R, \beta) , 
 \end{aligned}
 \right \} \label{eq:RVM_like_estimate_ALPHAs} 
\end{align}
The solution of \eqref{eq:LMMSE_estimate_X} is the standard linear minimum mean square error estimator (LMMSE) as
\begin{align}
&\mathrm{vec}(\hat{\mathbf{X}}) \leftarrow \beta \boldsymbol{\Sigma} \mathbf{A^\top  y} , \,\, \mathrm{where} \,\, \boldsymbol{\Sigma} = \left( (\boldsymbol{\alpha}_R \otimes \boldsymbol{\alpha}_L) +  \beta \mathbf{A^\top A} \right)^{-1}.
\nonumber
\end{align}
Using a standard approach (see equations (45) and (46) of \cite{Tipping1} or (7.88) of \cite{Bishop}), the solution of \eqref{eq:RVM_like_estimate_beta} can be found as
\begin{align}
\beta \leftarrow \frac{m + 2a}{||\mathbf{y} -\mathbf{A} \mathrm{vec}(\hat{\mathbf{X}})||_2^2 + \mathrm{tr}(\mathbf{A}\boldsymbol{\Sigma}\mathbf{A}^\top) + 2b}.
\label{eq:noise_update1}
\end{align}
The standard RVM in \cite{Tipping1} uses the different update rule \cite{mackay1991}
\begin{align}
\beta \leftarrow \frac{\mathrm{tr}(( \boldsymbol{\alpha}_R \otimes \boldsymbol{\alpha}_L)\boldsymbol{\Sigma}) + 2a}{||\mathbf{y} -\mathbf{A} \mathrm{vec}(\hat{\mathbf{X}})||_2^2 + 2b},
 \label{eq:noise_update2}
\end{align}
which often improves convergence \cite{mackay1991}. The update rule \eqref{eq:noise_update1} has the benefit over \eqref{eq:noise_update2} of having established convergence properties.
In simulations we used the update rule \eqref{eq:noise_update1} since it improved the estimation accuracy.

Finally we deal with \eqref{eq:RVM_like_estimate_ALPHAs} as follows.
\begin{enumerate}
 \item{\emph{For Schatten $s$-norm:}} Using the Schatten $s$-norm prior \eqref{eq:schatten_latent} gives us the update equations
\begin{align}
\begin{aligned}
&\boldsymbol{\alpha}_L \leftarrow c_s \left( \hat{\mathbf{X}}\boldsymbol{\alpha}_R\hat{\mathbf{X}}^\top + \tilde{\boldsymbol{\Sigma}}_L + \epsilon\mathbf{I}_p \right)^{(s-2)/2} ,\\
&\boldsymbol{\alpha}_R \leftarrow c_s \left( \hat{\mathbf{X}}^\top\boldsymbol{\alpha}_L\hat{\mathbf{X}} + \tilde{\boldsymbol{\Sigma}}_R + \epsilon\mathbf{I}_q \right)^{(s-2)/2} , 
\end{aligned}\label{eq:alpha_schatten}
\end{align} 
where $c_s = (s/2)^{s/2}$ and the matrices $\tilde{\boldsymbol{\Sigma}}_L$ and $\tilde{\boldsymbol{\Sigma}}_R$ have elements
\begin{align*}
&[\tilde{\boldsymbol{\Sigma}}_L ]_{ij} = \mathrm{tr}(\boldsymbol{\Sigma}(\boldsymbol{\alpha}_R \otimes \mathbf{E}_{ij}^{(L)})) , \\
&[\tilde{\boldsymbol{\Sigma}}_R]_{ij} = \mathrm{tr}(\boldsymbol{\Sigma}(\mathbf{E}_{ij}^{(R)} \otimes \boldsymbol{\alpha}_L )) , 
\end{align*}
and where $\mathbf{E}_{ij}^{(L)} \in \mathbb{R}^{p\times p}$ and $\mathbf{E}_{ij}^{(R)} \in \mathbb{R}^{q\times q}$ are matrices with ones in position $(i,j)$ and zeros otherwise.
 
 \item{\emph{Log-determinant penalty:}} For the log-determinant prior \eqref{eq:rsvm_latent} the update equations become
\begin{align}
\begin{aligned}
&\boldsymbol{\alpha}_L \leftarrow \nu  \left( \hat{\mathbf{X}}\boldsymbol{\alpha}_R\hat{\mathbf{X}}^\top + \tilde{\boldsymbol{\Sigma}}_L + \epsilon\mathbf{I}_p \right)^{-1} ,\\
&\boldsymbol{\alpha}_R \leftarrow \nu \left( \hat{\mathbf{X}}^\top\boldsymbol{\alpha}_L\hat{\mathbf{X}} + \tilde{\boldsymbol{\Sigma}}_R + \epsilon\mathbf{I}_q \right)^{-1}.
\end{aligned}  \label{eq:alpha_rsvm}
\end{align} 
We see that the update rule for the log-determinant penalty can be interpreted as \eqref{eq:alpha_schatten} in the limit $s \to 0$.
\end{enumerate}
The derivations of \eqref{eq:alpha_schatten} and \eqref{eq:alpha_rsvm} are shown in Appendix \ref{appendix2} and \ref{appendix3}. The corresponding update equations for the one-sided precision based model \eqref{eq:X_oneprior_1} are obtained by fixing the other precision matrix to be the identity matrix.
In the spirit of the evidence approximation based relevance vector machine, we call the developed algorithms in this section as relevance singular vector machine (RSVM). For Schatten-s norm and log-determinant priors, the methods are named as RSVM-SN and RSVM-LD, respectively.

\subsection{EM}

In expectation-maximization \cite{Bishop}, the value of the precisions $\theta \triangleq \{ \boldsymbol{\alpha}_L, \, \boldsymbol{\alpha}_R ,\, \beta\}$ are updated in each iteration by maximizing the cost (EM help function in MAP estimation)
\begin{align}
\label{eq:EM_cost}
 Q(\theta,\theta') + \log p(\theta)
\end{align}
where $\theta'$ are the parameter values from the previous iteration. The function $Q(\theta,\theta')$ is defined as
\begin{align}
& Q(\theta,\theta') \nonumber = \mathcal{E}_{\mathbf{X}|\mathbf{y},\theta'} [ \log p(\mathbf{y,X}|\theta)]  = \text{constant} \nonumber \\
& - \frac{\beta}{2} ||\mathbf{y - A}\mathrm{vec}(\hat{\mathbf{X}}) ||_2^2 - \frac{1}{2} \mathrm{tr}(\boldsymbol{\alpha}_L \hat{\mathbf{X}} \boldsymbol{\alpha}_R \hat{\mathbf{X}}^\top) -\frac{1}{2} \mathrm{tr}(\boldsymbol{\Sigma}^{-1} \boldsymbol{\Sigma}') \nonumber \\
& + \frac{q}{2} \log |\boldsymbol{\alpha}_L| + \frac{p}{2} \log |\boldsymbol{\alpha}_R| + \frac{m}{2} \log \beta,
\label{eq:LowRank:EM_help_function}
\end{align}
where $\boldsymbol{\Sigma}' = \left( (\boldsymbol{\alpha}'_R \otimes \boldsymbol{\alpha}'_L) + \beta' \mathbf{A^\top A} \right)^{-1}$, and $\mathcal{E}$ denotes the expectation operator. 
The maximization of $Q(\theta,\theta') + \log p(\theta)$ leads to update equations which are identical to the update equations of evidence approximation.
That means that for the Schatten-s norm, the maximization leads to \eqref{eq:LMMSE_estimate_X}, \eqref{eq:noise_update1} and \eqref{eq:alpha_schatten}, and for log-determinant penalty, the maximization leads to \eqref{eq:LMMSE_estimate_X}, \eqref{eq:noise_update1} and \eqref{eq:alpha_rsvm}. For the noise precision, EM reproduces the update equation \eqref{eq:noise_update1}.
The derivation of \eqref{eq:LowRank:EM_help_function} and update equations are shown in Appendix \ref{appendix:EM}.
Unlike evidence approximation, EM has monotonic convergence properties and hence the derived update equations are bound to improve estimation performance in iterations.

\subsection{Balancing the precisions}

We have found that in practical algorithms, there is a chance that one of the two precisions becomes large and the other small over iterations. 
A small precision results in numerical instability in the Kronecker covariance structure \eqref{eq:X_doubleprior}.
To prevent the inbalance we rescale the matrix precisions in each iteration such that $1)$ the a-priori and a-posteriori squared Frobeniun norm of $\mathbf{X}$ are equal,
\begin{align*}
&\mathcal{E}[||\mathbf{X}||_F^2 | \boldsymbol{\alpha}_L, \boldsymbol{\alpha}_R] = \mathrm{tr}(\boldsymbol{\alpha}_L^{-1}) \mathrm{tr}(\boldsymbol{\alpha}_R^{-1}) \\
&= \mathcal{E}[||\mathbf{X}||_F^2 | \boldsymbol{\alpha}_L, \boldsymbol{\alpha}_R,\beta, \mathbf{y} ] = ||\hat{\mathbf{X}}||_F^2 + \mathrm{tr}(\boldsymbol{\Sigma}),
\end{align*}
and $2)$ the contribution of the precisions to the norm is equal,
\begin{align*}
\mathrm{tr}(\boldsymbol{\alpha}_L^{-1}) = \mathrm{tr}(\boldsymbol{\alpha}_R^{-1}) .
\end{align*}
The rescaling makes the algorithm more stable and often improves estimation performance.

\section{Simulation experiments}
\label{sec:simulations}

In this section we numerically verify our two hypotheses,  and compare the new algorithms with relevant existing algorithms. Our objectives are to verify:
\begin{itemize}
 \item the hypothesis that the left-sided precision is better than the right sided precision for a fat low-rank matrix, 
 \item the hypothesis that the two-sided precision based model performs better than one-sided precision based model.
 \item the proposed methods perform better than a nuclear-norm minimization based convex algorithm and a variational Bayes algorithm. 
\end{itemize}
In the simulations we considered low-rank matrix reconstruction and also matrix completion as a special case due to its popularity.

\subsection{Performance measure, experimental setup and competing algorithms}

To compare the algorithms, the performance measure is the normalized-mean-square-error
\begin{align*}
\text{NMSE} \triangleq \mathcal{E}[||\hat{\mathbf{X}} - \mathbf{X}||_F^2]/\mathcal{E}[||\mathbf{X}||_F^2].
\end{align*}

In experiments we varied the value of one parameter while keeping the other parameters fixed. For given parameter values, we evaluated the NMSE as follows.

\begin{enumerate}

\item For LRMR, the random measurement matrix $\mathbf{A} \in \mathbb{R}^{m \times pq}$ was generated by independently drawing the elements from $\mathcal{N}(0,1)$ and normalizing the column vectors to unit norm. For low rank matrix completion, each row of $\mathbf{A}$ contains a 1 in a random position and zero otherwise with the constraint that the rows are linearly independent.

\item Matrices  $\mathbf{L} \in \mathbb{R}^{p \times r}$ and $\mathbf{R} \in \mathbb{R}^{r \times q}$ with elements drawn from $\mathcal{N}(0,1)$ were randomly generated and the matrix $\mathbf{X}$ was formed as $\mathbf{X = LR}$. Note that $\mathbf{X}$ is of rank $r$ (with probability $1$). 

\item Generate the measurement $\mathbf{y} = \mathbf{A} \mathrm{vec}(\mathbf{X}) + \mathbf{n}$, where $\mathbf{n} \sim \mathcal{N}(\mathbf{0},\sigma_n^2 \mathbf{I}_m)$ and $\sigma_n^2$ is chosen such that the signal-to-measurement-noise ratio is
\begin{align*}
\mathrm{SMNR} \triangleq \frac{\mathcal{E}[||\mathbf{A} \mathrm{vec}(\mathbf{X})||_2^2]}{\mathcal{E}[||\mathbf{n}||_2^2]} = \frac{rpq}{m\sigma_n^2}.
\end{align*}

\item Estimate $\hat{\mathbf{X}}$ using competing algorithms and calculate the error $||\hat{\mathbf{X}} - \mathbf{X}||_F^2$. 

\item Repeat steps $2-4$ for each measurement matrix $T_1$ number of times.

\item Repeat steps $1-5$ for the same parameter values $T_2$ number of times.

\item Then compute the NMSE by averaging.

\end{enumerate}

In the simulations we chose $T_1 = T_2 = 25$, which means that the averaging was done over 625 realizations. We normalized the column vectors of $\mathbf{A}$ to make the SMNR expression realization independent.

Finally we describe competing algorithms.
For comparison, we used the following nuclear norm based estimator
\begin{align*}
\hat{\mathbf{X}} = \arg \min_\mathbf{X} ||\mathbf{X}||_* \text{ , s.t. } ||\mathbf{y - A}\mathrm{vec}(\mathbf{X})||_2 \leq \epsilon,
\end{align*}
where we used $\epsilon = \sigma_n \sqrt{m + \sqrt{8m}}$ as proposed in \cite{candes2006}. The cvx toolbox \cite{cvx} was used to implement the estimator. 
For matrix completion we also compared with the Variational Bayesian (VB) developed by Babacan et. al. \cite{Babacan}. In VB, the matrix $\mathbf{X}$ is factorized as
\begin{align*}
\mathbf{X = \underbar{L} \, \underbar{R}},
\end{align*}
and (block) sparsity inducing priors are used for the column vectors of $\mathbf{\underbar{L}} \in \mathbb{R}^{p \times \min(p,q)}$ and $\mathbf{\underbar{R}}^\top \in \mathbb{R}^{q \times \min(p,q)}$. The VB algorithm was developed for matrix completion (and robust PCA), but not for matrix reconstruction. We note that, unlike RSVM and VB, the nuclear norm estimator requires a-priori knowledge of the noise power. 
We also compared the algorithms to the Cram{\'e}r-Rao bound (CRB) from \cite{Werner,Tang2011} (as we know the rank a-priori in our experimental setup). 
We mention that the CRB is not always a valid lower bound in this experimental setup because all technical conditions for computing a valid CRB are not always fulfilled and the estimators are not always unbiased. The choice of CRB is due to absence of any other relevant theoretical bound.

\subsection{Simulation results}

\begin{figure}[t]
\begin{center}
\includegraphics[width=0.9\linewidth]{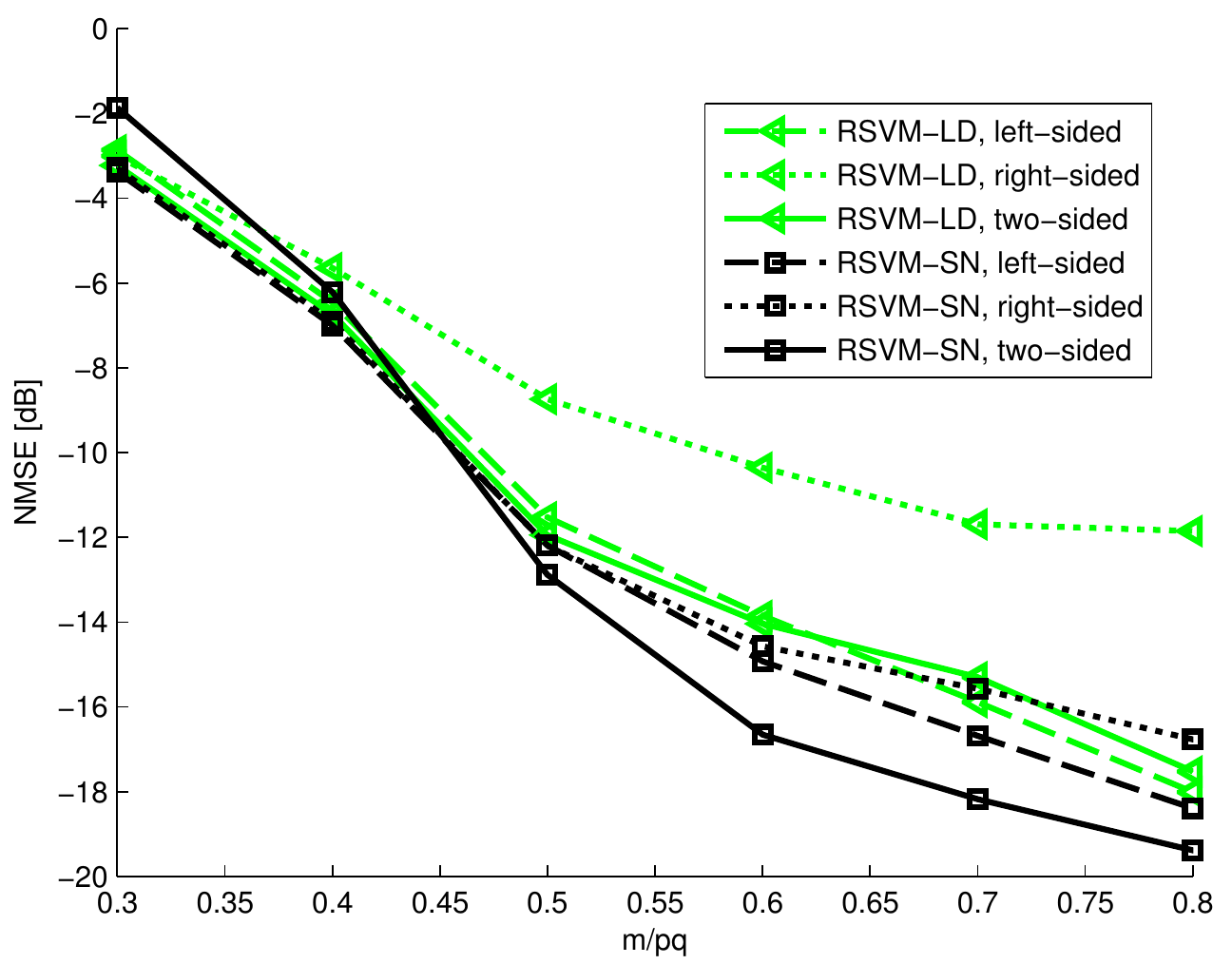}
\caption{NMSE vs. $m/(pq)$ for low-rank matrix reconstruction.}
\label{fig:single_double_rec}
\end{center}
\end{figure}

Our first experiment is for verification of the first two hypotheses. 
For the experiment, we considered LRMR and fixed $\mathrm{rank}(\mathbf{X}) = r = 3$, $p = 15$, $q = 30$, $\mathrm{SMNR} = 20$ dB and 
varied $m$. The results are shown in Figure~\ref{fig:single_double_rec}
where NMSE is plotted against normalized measurements $m/(pq)$.
We note that RSVM-SN with left precision is better than right precision. Same result also hold for RSVM-LD. 
This verifies the first hypothesis. Further we see that RSVM-SN and RSVM-LD with two sided precisions
are better than respective one-sided precisions. This result verifies the second hypothesis.
In the experiments we used $s=0.5$ for RSVM-SN as it was found to be the best (empirically).
Henceforth we fix $s=0.5$ for RSVM-SN.

\begin{figure}[t]
\begin{center}
\includegraphics[width=0.9\linewidth]{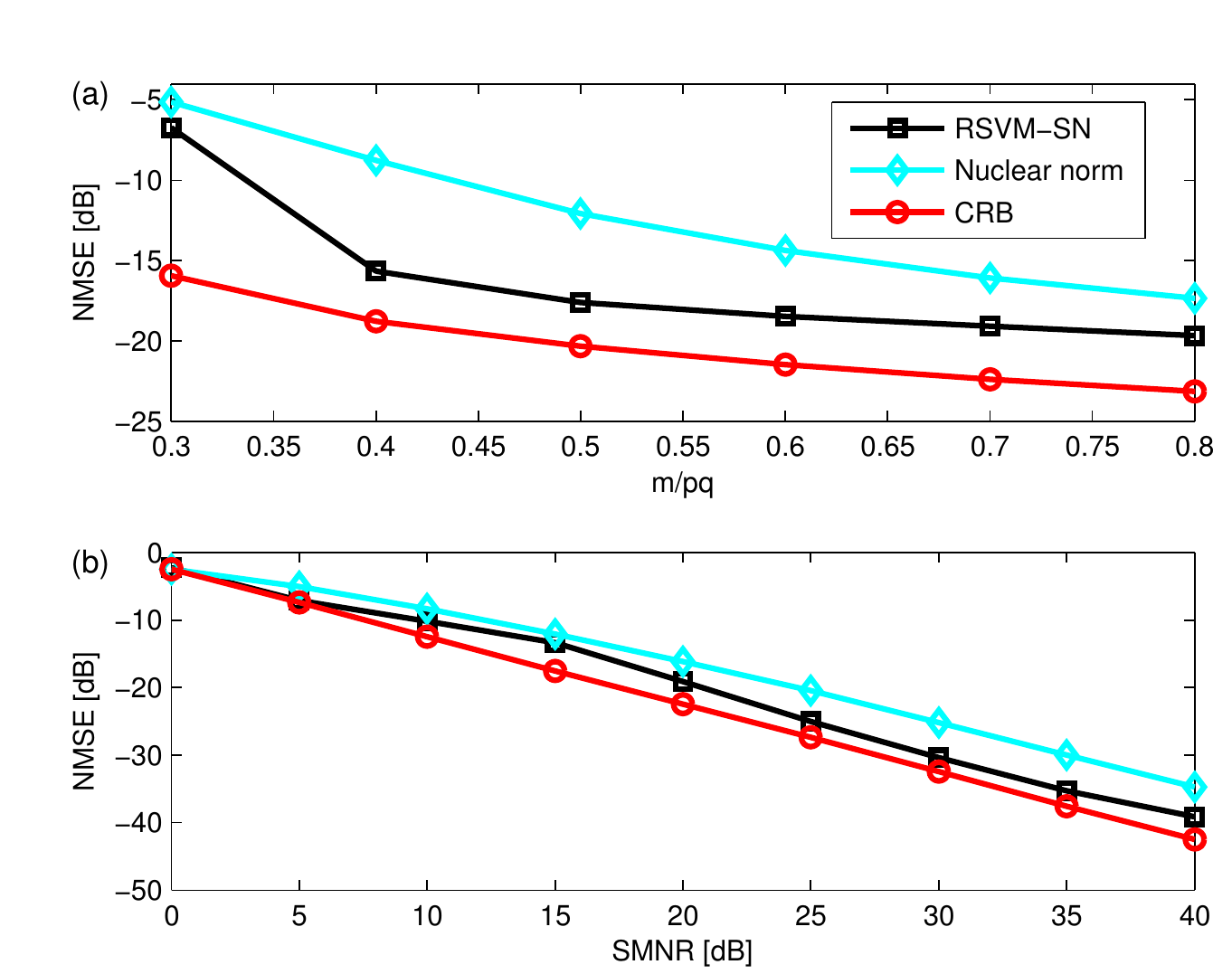}
\caption{NMSE vs. $m/(pq)$ and SMNR for low-rank matrix reconstruction. (a) SMNR = 20 dB and $m/(pq)$ is varied. (b) $m/(pq)=0.7$ and SMNR is varied.}
\label{fig:alpha_snr_rec}
\end{center}
\end{figure}

The second experiment considers comparison with nuclear-norm based algorithm and the CRB for LRMR.
The objective is robustness study by varying number of measurements and measurement noise power.
We used $r = 3$, $p = 15$ and $q = 30$. In Figure~\ref{fig:alpha_snr_rec} (a) we show
the performance against varying $m/(pq)$; the SMNR = 20 dB was fixed. 
The performance improvement of RSVM-SN is more pronounced over the nuclear-norm based algorithm
in the low measurement region. Now we fix $m/(pq)=0.7$ and vary the SMNR. The results are shown in Figure~\ref{fig:alpha_snr_rec} (b)
which confirms robustness against measurement noise.

\begin{figure}[t]
\begin{center}
\includegraphics[width=0.9\linewidth]{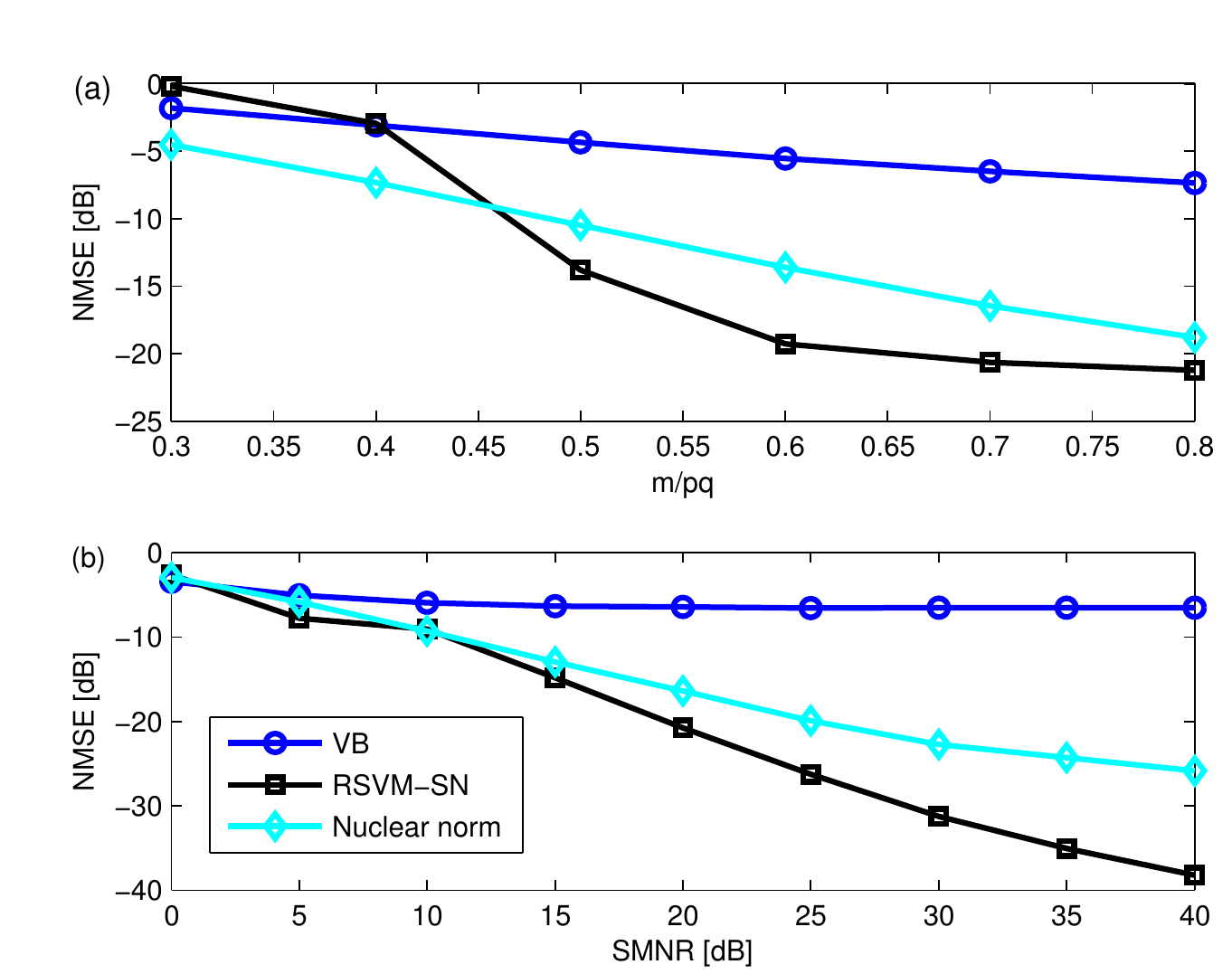}
\caption{NMSE vs. $m/(pq)$ and SMNR for low-rank matrix completion. (a) SMNR = 20 dB and $m/(pq)$ is varied. (b) $m/(pq)=0.7$ and SMNR is varied.}
\label{fig:alpha_snr_comp}
\end{center}
\end{figure}

Next we deal with matrix completion where the measurement matrix $\mathbf{A}$ has a special structure
and considered to be inferior to hold information about $\mathbf{X}$ than the same dimensional random measurement matrix used in LRMR.
Therefore matrix completion requires more measurements and higher SMNR.
We performed similar experiments as in our second experiment and the results are shown in Figure~\ref{fig:alpha_snr_comp}.
In the experiments the performance of the VB algorithm is included. It can be seen that RSVM-SN is typically better than the other algorithms.
We find that the the VB algorithm is pessimistic.

\begin{figure}[t]
\begin{center}
\includegraphics[width=0.9\linewidth]{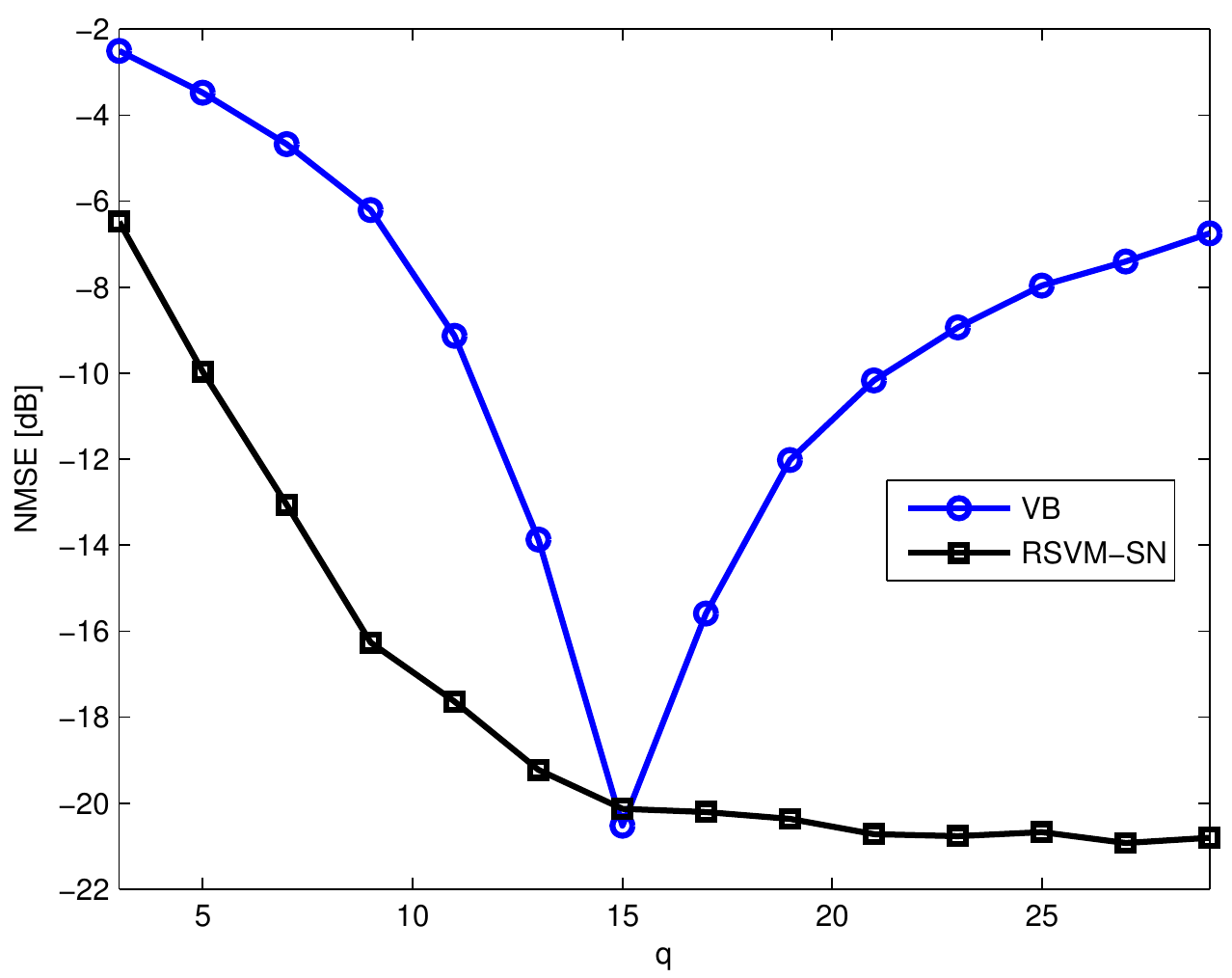}
\caption{NMSE vs. $q$ for low-rank matrix completion.}
\label{fig:q_completion}
\end{center}
\end{figure}

Finally in our last experiment we investigated the VB algorithm to find conditions for
its improvement and compared it with RSVM-SN. For this experiment, we fixed $r = 3$, $p = 15$, $m/(pq)=0.7$ and SMNR = 20 dB, and varied $q$.
The results are shown in Figure~\ref{fig:q_completion} and we see that VB provides good performance when $p=q$. The result may be attributed to an aspect that
VB is highly prone to a large number of model parameters which arises in case $\mathbf{X}$ is away from a square matrix.

\section{Conclusion}

In this paper we developed Bayesian learning algorithms for low-rank matrix reconstruction. The framework relates low-rank penalty functions (type I estimators) to the latent variable models (type II estimators) with either left- or right-sided precisions through the matrix Laplace transform and the concave conjugate formula. The model was further extended to the two-sided precision based model. Using evidence approximation and expectation maximization, we derived the update equations for the parameters. The resulting algorithm was named the Relevance Singular Vector Machine (RSVM) due to its similarity with the Relevance Vector Machine for sparse vectors. Especially we derived the update equations for the estimators corresponding to the log-determinant penalty and the Schatten $s$-norm penalty, we named the algorithms RSVM-LD and RSVM-SN, respectively.

Through simulations, we showed that the two-sided precision based model performs better than the one-sided model for matrix reconstruction. The algorithm also outperformed a nuclear-norm based estimator, even though the nuclear-norm based estimator knew the noise power. The proposed methods also outperformed a variational Bayes method for matrix completion when the matrix is not square.

\appendix[Derivations]

\subsection{Derivation of the Laplace Approximation}

\label{appendix:laplace}

The Laplace approximation is an approximation of the integral
\begin{align*}
I = \int e^{-\frac{1}{2} f(\mathbf{a})} d\mathbf{a},
\end{align*}
where the integral is over $\mathbf{a} \in \mathbb{R}^n$. The function $f(\mathbf{a})$ is approximated by a second order polynomial around its minima $\mathbf{a}_0$ as
\begin{align*}
f(\mathbf{a}) \approx f(\mathbf{a}_0) + \frac{1}{2} (\mathbf{a - a}_0)^\top \mathbf{H} (\mathbf{a - a}_0),
\end{align*}
where $\mathbf{H} = \nabla^2 f(\mathbf{a}) |_{\mathbf{a = a}_0}$ is the Hessian of $f(\mathbf{a})$ at $\mathbf{a}_0$. The term linear in $\mathbf{a}$ vanishes and $\mathbf{H} \succ \mathbf{0}$ at $\mathbf{a}_0$ since we expand around a minima. With this approximation, the integral becomes
\begin{align*}
I \approx \int e^{-\frac{1}{2} f(\mathbf{a}_0) - \frac{1}{4} (\mathbf{a - a}_0)^\top \mathbf{H} (\mathbf{a - a}_0)} d\mathbf{a} = \sqrt{\frac{(4\pi)^{n}}{|\mathbf{H}|}} e^{-\frac{1}{2} f(\mathbf{a}_0)}.
\end{align*}
In \eqref{eq:marginal_x}, the integral is given by
\begin{align*}
I = \frac{1}{(2\pi)^{pq/2}} \int_{\boldsymbol{\alpha} \succ \mathbf{0}} e^{-\frac{1}{2} [ \mathrm{tr}(\boldsymbol{\alpha}\mathbf{Z}) -q\log|\boldsymbol{\alpha}|  + K(\boldsymbol{\alpha}) ]} d \boldsymbol{\alpha}.
\end{align*}
Set $f(\mathbf{a}) = \mathrm{tr}(\boldsymbol{\alpha}\mathbf{Z}) -q\log|\boldsymbol{\alpha}|  + K(\boldsymbol{\alpha})$, where $\mathbf{a} = \mathrm{vec}(\boldsymbol{\alpha})$. Let $\boldsymbol{\alpha}_0 \succ \mathbf{0}$ denote the minima of $f(\mathbf{a})$ and $\mathbf{H}$ the Hessian at $\boldsymbol{\alpha}_0$. Assuming that $\boldsymbol{\alpha}_0$ and $\mathbf{H}$ are ``large'' in the sense that the integral over $\boldsymbol{\alpha} \succ \mathbf{0}$ can be approximated by the integral over $\boldsymbol{\alpha} \in \mathbb{R}^{p \times p}$ we find that
\begin{align*}
&I \approx \frac{1}{(2\pi)^{pq/2}} \int e^{-\frac{1}{2} f(\mathbf{a}_0) - \frac{1}{4} (\mathbf{a - a}_0)^\top \mathbf{H} (\mathbf{a - a}_0)} d\mathbf{a} \\
&= \frac{(4\pi)^{p^2/2}}{(2\pi)^{pq/2} |\mathbf{H}|^{1/2}} e^{-\frac{1}{2} f(\mathbf{a}_0)},
\end{align*}
where $\mathbf{a}_0 = \mathrm{vec}(\boldsymbol{\alpha}_0)$.

\subsection{The EM help function}

\label{appendix:EM}

The EM help function $Q(\theta,\theta')$ is given by
\begin{align*}
&Q(\theta,\theta')  = \mathcal{E}_{\mathbf{X}| \mathbf{y},\theta'} [\log \, p(\mathbf{X}|\mathbf{y},\theta)] = c + \frac{m}{2} \log \, \beta \\
&- \frac{\beta}{2} \mathcal{E}[||\mathbf{y - A}\mathrm{vec}(\mathbf{X})||_2^2 - \frac{1}{2} \mathcal{E}[\mathrm{tr}(\boldsymbol{\alpha}_L \mathbf{X} \boldsymbol{\alpha}_R \mathbf{X}^\top)] \\
&+ \frac{q}{2} \log |\boldsymbol{\alpha}_L| + \frac{p}{2} \log |\boldsymbol{\alpha}_R|,
\end{align*}
where $c$ is a constant. Using that
\begin{align*}
&\mathcal{E}[||\mathbf{y - A}\mathrm{vec}(\mathbf{X})||_2^2] = ||\mathbf{y}||_2^2 - 2\mathbf{y}^\top \mathbf{A} \mathrm{vec}(\hat{\mathbf{X}}) \\
&+ \mathrm{tr}(\mathbf{A^\top A}(\mathrm{vec}(\hat{\mathbf{X}}) \mathrm{vec}(\hat{\mathbf{X}})^\top + \boldsymbol{\Sigma}')) \\
&= ||\mathbf{y - A}\mathrm{vec}(\hat{\mathbf{X}})||_2^2 + \mathrm{tr}(\mathbf{A}^\top\mathbf{A}\boldsymbol{\Sigma}') , 
\end{align*}
and
\begin{align*}
&\mathcal{E}[\mathrm{tr}(\boldsymbol{\alpha}_L \mathbf{X} \boldsymbol{\alpha}_R \mathbf{X}^\top)] \\
&= \mathrm{tr}((\boldsymbol{\alpha}_R \otimes \boldsymbol{\alpha}_L)(\mathrm{vec}(\hat{\mathbf{X}}) \mathrm{vec}(\hat{\mathbf{X}})^\top + \boldsymbol{\Sigma}')) \\
&= \mathrm{tr}(\boldsymbol{\alpha}_L \hat{\mathbf{X}} \boldsymbol{\alpha}_R \hat{\mathbf{X}}^\top) + \mathrm{tr}((\boldsymbol{\alpha}_R \otimes \boldsymbol{\alpha}_L) \boldsymbol{\Sigma}'),
\end{align*}
we recover the expression \eqref{eq:LowRank:EM_help_function} for the EM help function.

\subsection{Details for the RSVM with the Schatten $s$-norm penalty}

\label{appendix2}

We here set $\mathbf{S} = \epsilon \mathbf{I}_q$ to keep the derivation more general. The regularized Schatten $s$-norm penalty is given by
\begin{align*}
\tilde{g}(\mathbf{Z}) = \mathrm{tr}((\mathbf{X^\top X} + \mathbf{S})^{s/2}).
\end{align*}
For the concave conjugate formula \eqref{eq:inverse_conjugate} we find that the minimum over $\mathbf{Z}$ occurs when
\begin{align*}
\boldsymbol{\alpha} - \frac{s}{2} (\mathbf{Z + S})^{s/2-1} = \mathbf{0}.
\end{align*}
Solving for $\mathbf{Z}$ gives us that
\begin{align*}
\tilde{K}(\boldsymbol{\alpha}) = - \mathrm{tr}(\boldsymbol{\alpha}\mathbf{S}) - \frac{2-s}{s} \left( \frac{2}{s} \right)^{-2/(2-s)} \mathrm{tr}(\boldsymbol{\alpha}^{-2/(2-s)}),
\end{align*} 
which results in \eqref{eq:schatten_latent}.

Using \eqref{eq:LowRank:EM_help_function}, we find that the minimum of \eqref{eq:EM_cost} for the Schatten $s$-norm occurs when
\begin{align*}
\hat{\mathbf{X}}\boldsymbol{\alpha}_R \hat{\mathbf{X}}^\top + \tilde{\boldsymbol{\Sigma}}_R  - \left( \frac{2}{s} \right)^{-s/(2-s)} \boldsymbol{\alpha}_L^{-2/(2-s)} = \mathbf{0}
\end{align*}
Solving for $\boldsymbol{\alpha}_L$ gives \eqref{eq:alpha_schatten} for $\boldsymbol{\alpha}_L$. The update equation for $\boldsymbol{\alpha}_R$ is derived in a similar manner.

\subsection{Details for the RSVM with the log-determinant penalty}

\label{appendix3}

The log-determinant penalty is given by
\begin{align*}
g(\mathbf{X}) = \nu \log | \mathbf{Z + S} |.
\end{align*}
For the concave conjugate formula \eqref{eq:inverse_conjugate} we find that the minimum over $\mathbf{Z}$ occurs when
\begin{align*}
\boldsymbol{\alpha} - \nu (\mathbf{Z + S})^{-1} = \mathbf{0}.
\end{align*}
Solving for $\mathbf{Z}$ gives
\begin{align*}
\tilde{K}(\boldsymbol{\alpha}) = - \mathrm{tr}(\boldsymbol{\alpha}\mathbf{S}) + \nu \log |\boldsymbol{\alpha}| + \nu p - \nu \log \nu.
\end{align*} 
By removing the constants we recover \eqref{eq:rsvm_latent}.

Using \eqref{eq:LowRank:EM_help_function}, we find that the minimum of \eqref{eq:EM_cost} with respect to $\boldsymbol{\alpha}_L$ for the log-determinant penalty occurs when
\begin{align*}
\hat{\mathbf{X}}\boldsymbol{\alpha}_R \hat{\mathbf{X}}^\top + \tilde{\boldsymbol{\Sigma}}_R  + \mathbf{S}_L - \nu \boldsymbol{\alpha}_L^{-1} = \mathbf{0}
\end{align*}
Solving for $\boldsymbol{\alpha}_L$ gives us \eqref{eq:alpha_rsvm} for $\boldsymbol{\alpha}_L$. The derivation of the update equation for $\boldsymbol{\alpha}_R$ is found in a similar way.

\end{document}